# Combining transmission speckle photography and convolutional neural network for determination of fat content in cow's milk - an exercise in classification of parameters of a complex suspension

Kwasi Nyandey[a,b,*] and Daniel Jakubczyk[a]

[a]Institute of Physics, Polish Academy of Sciences, Al. Lotników 32/46, 02-668 Warsaw, Poland

[b]Laser and Fibre Optics Centre, Department of Physics, School of Physical Sciences, College of Agriculture and Natural Sciences, University of Cape Coast, Cape Coast, Ghana

*Corresponding author: kwasi.nyandey@ucc.edu.gh

**Abstract**

We have combined transmission speckle photography and machine learning for direct classification and recognition of milk fat content classes. Our aim was hinged on the fact that parameters of scattering particles (and the dispersion medium) can be linked to the intensity distribution (speckle) observed when coherent light is transmitted through a scattering medium. For milk, it is primarily the size distribution and concentration of fat globules, which constitutes the total fat content. Consequently, we trained convolutional neural network to recognise and classify laser speckle from different fat content classes (0.5, 1.5, 2.0 and 3.2%). We investigated four exposure-time protocols and obtained the highest performance for shorter exposure times, in which the intensity histograms are kept similar for all images and the most probable intensity in the speckle pattern is close to zero. Our neural network was able to recognize the milk fat content classes unambiguously and we obtained the highest test and independent classification accuracies of 100 and ~99% respectively. It indicates that the parameters of other complex realistic suspensions could be classified with similar methods.

## 1. Introduction

In recent years the demand for fast, simple and non-destructive determination of selected properties of food products is increasing, as the socially expected quality standards are increasing worldwide. Consequently, there have been several developments to couple the existing (primarily optical) experimental methods with recent technological advancements to tackle this challenge. The reports in literature include e.g. the combination of spectroscopy and machine learning to measure selected constituents of milk [1,2], and the use of spectroscopic and chemometric techniques to determine quality parameters in intact pineapple fruits [3] as well as to non-destructively determine the freshness of eggs [4]. Milk, in particular, is known to be an important and nutrient-rich liquid food product and also a naturally occurring suspension with diverse constituents. The methods developed for milk should be easily adaptable to other real-world suspensions.

In raw milk, the constituents of interest are (beside water) protein (~3.4 wt%), lipids (fat; ~3.6 wt%) and lactose (~4.6 wt%). The dominant component of total protein in raw cow milk is casein micelle (about 80wt%), where $\alpha_1$-, $\alpha_2$- and $\beta$-casein molecules – in approximately constant ratios – combined with calcium phosphate to form aggregates whose surfaces are stabilised by $\kappa$-casein molecules [5]. It restricts further growth of the aggregates to form precipitates. Consequently, the size variability ranges between 50 and 680 nm [6]. On the other hand, fat globules in raw milk are known to be secreted as droplets (covered by a layer of protein acting as an emulsifier) of variable sizes (1 – 10 μm) [7,8] and in different amounts. Also, lactose is the naturally occurring sugar in milk, which is dissolved in the serum phase of milk. It has two anomers: $\alpha$-lactose and $\beta$-lactose, whose proportion depends mainly on the temperature [9]. The concentration, composition and structure of protein, fat and lactose in milk have been the main target of several optical research [2,8,10], replacing their chemical counterparts [6,7]. The quality of milk is measured from its protein, fat and lactose content [7,11] and other constituents of less practical applicability.

For commercial and consumption purposes milk is usually processed and labelled according to their fat and/or lactose contents. This processing involves several stages of treatments (and testing) which is known to cause significant reduction in the concentrations and sizes of milk constituents [8]. Hence a simple and straightforward technique (which can be made online) will ensure accurate and fast measurement of these nutritional parameters. Such information will be beneficial for consumer protection, industrial and engineering purposes as well as for best cow farm management practices. As a matter of fact, several attempts have already been made. For instance, several optical (ranging from UV to IR) methods – exploiting several phenomena – have been established to estimate the content and composition of protein, fat and lactose, as reviewed by [10]. However, almost all of these methods are based on the acquisition of spectral fingerprints of such selected constituents of milk, requiring (in some cases) sophisticated experimental setup as well as chemical sample preparation. These methods are mostly offline which is not always applicable to routine everyday use. A different method was proposed in [12], whereby a conventional digital camera was used to record RGB LED light transmitted through 4-mm-thick milk samples at different exposure times. The fat and protein contents were successfully correlated with the intensity distributions observed in the images. Although the method has not been verified on independent samples, it can be considered very promising.

With several attempts still being made towards simple and non-destructive determination of milk nutritional parameters, we have investigated the potential of combining transmission speckle photography and machine learning for direct classification and recognition of milk fat content classes. For the sake of generality (and simplicity) we used off-shelf milk samples which are common and usually produced from a mixture of milk at different stages of lactation and from several cows. As hinted above, with this exercise, we have in mind developing a method of characterisation of complex suspensions.

It is widely accepted that coherent scattering of light carries detailed information about the scattering particles, but also more generally: random refractive index fluctuations. This interference phenomenon (speckle) is extremely sensitive to scattering medium properties and is an established subject matter as well as a versatile tool for a wide range of measurements [13–16] in industry, engineering, biology and even medicine. It has been exploited for measurements such as deformation [17], random processes at rough surfaces [18], as well as endoscopic application [19], and blood perfusion velocity prediction [20]. Mainly because these methods are non-contact and most often require little or no sample preparation. In modern times, digital and computer-assisted recording and evaluation of speckle patterns have further extended their applications. It should be admitted that for essentially inhomogeneous and poly-disperse suspension, such as milk, there is no known explicit relation between the speckle patterns and the properties of the suspension involved. However, there are several mature methods of extracting information from the speckle field (evolution).

When measurements in which the speckle is essentially static are concerned, the coordinated speckle movement is only possible after deformation or displacement of the sample. It is relatively easy then to evaluate the speckle pattern change and draw conclusions, e.g. with the phase shift technique (see e.g. [21]). In the case of suspension, such as milk, whose particles are in constant random motion, analysis of the speckle pattern becomes challenging [22]. Phase shift technique would be of little use, since the individual motion of the particles does not result in a common phase difference between reference and object beams but in a locally random one. Therefore, an established technique of the dynamic speckle analysis – dynamic light scattering (DLS) – is based on the analysis of temporal correlations of light intensity observed from a small spatially coherent domain. The method yields characteristic times of the system, so the obtained parameters of the studied suspension are somewhat model-dependent. It also requires detectors with near-nanosecond resolution and thus cannot be cheaply extended to image sequence analysis.

On the contrary, a direct relation of the fat content (size distribution and concentration of fat globules) to speckle pattern seems a better alternative. Fat content in UHT milk is still comparable to casein micelle content [23], while fat globules are significantly larger and thus constitute the main source of speckle (scattering). Then, machine learning (ML) can be used to classify patterns and retrieve the "encoded" information. A deep neural network is known to learn rich feature representations of a wide range of data including images [24] without the need for rigorous patterns analysis. It can be perceived that a convolutional neural network (CNN), such as we used in this study, performs some type of correlation, which is very much in line with the existing methods of speckle field analysis. Hence we propose a technique, which involves recording dynamic speckles at relatively low frame rate, 75 fps. The movie frames (treated independently in this experiment) serve as the speckle patterns. The individual speckle patterns then carry information about the scattering particles [13] – which are fat globules, protein micelles, etc., in the case of milk – as well as their sizes and/or concentrations. With recent GPU capabilities, advanced machine learning (ML) can be applied to recognise and classify the patterns into their appropriate fat content classes. Our earlier experiment with direct recognition of suspensions with different nanoparticle materials, sizes and concentrations in thin cuvette using speckle photography and ML [25] confirmed this possibility.

However, there are several issues that need to be clarified during the study. There is almost no report in the literature on optimal speckle properties, such as brightness distribution (camera exposure time) suitable for full speckle development, as well as feature extraction in a neural network. Due to the movement of scatterers, the intensity histogram – brightness level and distribution as well as other statistical parameters differ not only for each concentration but for every pattern rec-orded. Though manipulation of brightness level has been used as image augmentation technique in the machine learning environment to either enhance image quality or create several copies of a single image [26], it is still unclear (to best of our knowledge) how these parameters in dynamic speckle affect the performance of neural networks.

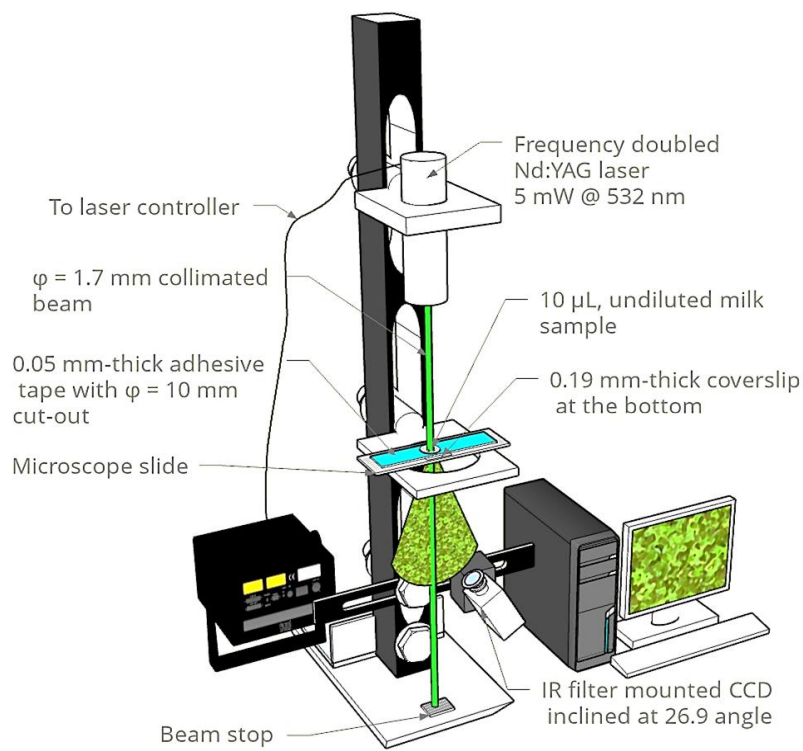

Figure 1: Experimental setup for recording speckle movies from the milk samples.

## 2. Experiment

### 2.1 *Experimental Setup*

A flat thin cuvette was prepared by sticking a 50 μm-thick adhesive tape on a microscope slide and cutting out a 10 mm circular segment from the centre of the tape. The undiluted milk sample (10 μL) was loaded into the cuvette and gently covered with a 0.19-mm-thick coverslip. By utilising a thin cuvette we minimised multiple scattering, thus avoiding non-linear effects in brightness. As represented in Figure 1, the sample stage was levelled so that the motion of fat globules will be purely Brownian. The sample was illuminated perpendicularly with a 5 mW, 1.7 mm collimated beam from a green frequency-doubled Nd:YAG laser at 532 nm. The selection of this source was determined because of the low absorption coefficient of milk within this wavelength range as reported in [27]. The illumination was such that the speckle exited through the thin coverslip to minimise static scattering in the glass. A 14-bit colour camera (Pike F-032C, AVT) was used to record the speckle at 26.9° angle, taking into account the velocity of the speckle movement and camera's available frame rate. Also an IR filter was mounted on the camera to account for the leak of the fundamental and pump frequencies from the laser source in infrared. The experiment was performed in a dark room at 21±1°C.

| Market label | Sample label | Fat content [wt%] | Number of lots | Milk type |
|---|---|---|---|---|
| Łaciate | Dairy 1 | 0.5 | 15 | UHT |
| | | 2.0 | 15 | UHT |
| | | 3.2 | 15 | UHT |
| Carrefour | Dairy 2 | 0.5 | 5 | UHT |
| | | 2.0 | 5 | UHT |
| | | 3.2 | 5 | UHT |
| Mleczna Dolina | Dairy 3 | 0.5 | 5 | UHT |
| | | 1.5 | 5 | UHT |
| | | 2.0 | 5 | Fresh |
| | | 3.2 | 5 | UHT |
| Mlekovita | Dairy 4 | 1.5 | 5 | UHT |
| Zambrowskie | Dairy 5 | 1.5 | 2 | UHT |
| Mixed Dairy | | 0.5 | $2_1, 2_2, 1_3$ | UHT |
| | | 1.5 | $2_3, 2_4, 1_5$ | UHT |
| | | 2.0 | $2_1, 2_2, 1_3$ | UHT & Fresh |
| | | 3.2 | $2_1, 2_2, 1_3$ | UHT |

Table 1: Distribution of cow milk types and fat content classes according to dairy plants. For Mixed Dairy, proportions of 2:2:1 lots were mixed – subscript denotes the diary number. Dairy 1 = Mlekpol ZPM Grajewo, Dairy 2 = OSM Wart-Milk Sieradz, Dairy 3 = OSM Łowicz, Dairy 4 = Mlekovita, Wysokie Mazowieckie, Dairy 5 = Mlekpol ZPM Zambrów.

### 2.2 *Sample collection*

The milk of different fat content classes was purchased from supermarkets (where fresh milk was refrigerated but ultra-pasteurized (UHT) milk was left on the open shelf). As certified by independent food technologists in Poland, milk composition data provided by the manufacturer could be depended upon. All the milk samples were homogenised, but the details of the procedure were not stated by the manufacturers. Perhaps, it should be reminded here that homogenization of milk only reduces the polydispersity of the fat globules, while leaving the milk an essentially inhomogeneous complex suspension.

As shown in Table 1, the samples were mainly from 3 dairy plants. Dairies 4 and 5 were added to supplement the lack of 1.5% class in Dairies 1 and 2. Though the Dairy 5 and Dairy 1 belong to the same company the dairy plants are different. The production rate of Dairy 5 was very low, hence sampling different lots was a challenge. However, with addition of these extra dairies it enabled us to create a separate training sample with mixed dairies. For the composition of the Mixed Dairy, we mixed images in proportions of 2:2:1 lots (see Table 1).

Initially we experimented with milk from one dairy plant (Dairy 1) and then confirmed our findings with milk from other dairy plants. Hence for each fat content class (0.5, 2.0 and 3.2%) produced by Dairy 1 we sampled 15 1L-containers of milk with different batch/lot numbers. For Dairies 2-4 we sampled 5 1L-containers and for Dairy 5 – 2 such containers. It must be stressed that we adhered to the rule that each 1L-container belonged to a different lot, hereafter referred to as different lots. We used milk from only freshly opened containers.

However, it must be kept in mind that there are seasonal variations of milk properties during the annual cycle – batches separated by an interval of several weeks may be different enough to introduce an additional parameter into the learning process. Constructing a fully representative dataset would then be quite lengthy. We tried to avoid that, since we could not afford to run the project for several months while we aimed only at checking the method's potential.

*2.3 Speckle recording procedures*

For each lot, two experimental runs were performed. In the first experimental run, the milk was loaded into the cuvette and 10 movies were recorded at 10 different locations on the sample (different background/static speckle) and labelled as "Version 1". After recording 10 movies for Version 1 the cuvette was cleaned. It involved gently dipping it in a diluted cleaning solution, sonication for a minute and rinsing off with distilled water. Finally it was dried clean by blowing dry nitrogen gas (99.8%) over it. It was then loaded again with milk to record the movies for Version 2. At a single illumination point, 13 s (75 fps) movie was recorded, the 10-locations on the same sample was exhaustive. Keeping the loaded sample in the cuvette for too long was avoided since it resulted in drying of the sample. We saved movies of each lot and each version in separate directories and af-ter extraction (as will be discussed in section 2.5) the images were named according to the movie label and frame number.

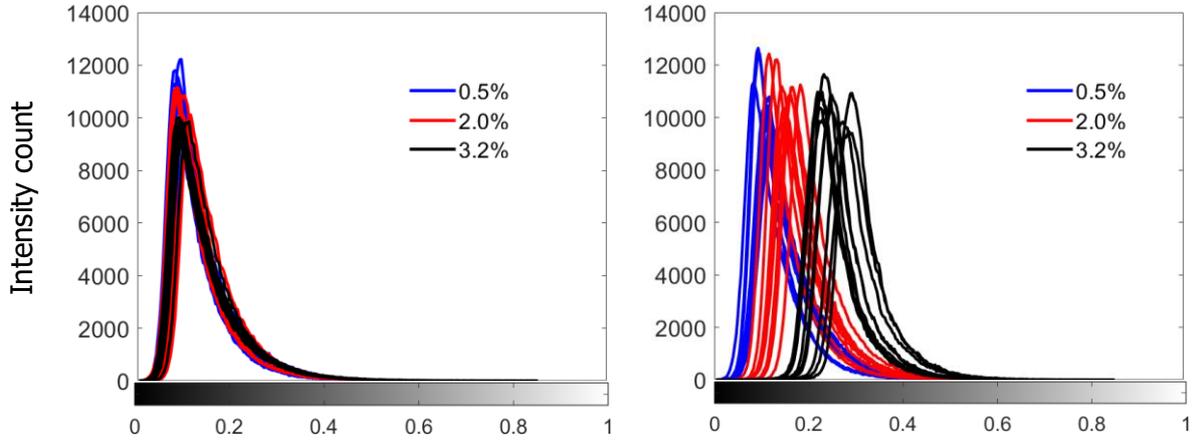

Figure 2: Intensity histograms of speckle patterns corresponding to the first 10 consecutive frames of movies for each sample type from Protocol A (left panel) and B (right panel).

**2.4 *Exposure analysis/protocols***

Generally, a fully developed speckle pattern is limited to an intensity band and obeys negative exponential probability distribution as presented by Goodman in [13]. The marginal probability density for a chosen intensity *I* is given by

$$p_I(I) = \frac{1}{2\sigma^2} \exp\left(-\frac{I}{2\sigma^2}\right) \text{ for } I \geq 0 \text{ and obviously 0 otherwise,} \quad (1)$$

where σ² is the variance of the complex field. Practically, it is desirable that the most probable intensity in the pattern is zero. In which case the probability that the intensity exceeds a threshold *I* is

$$P(I) = \exp\left(-\frac{I}{\langle I \rangle}\right). \quad (2)$$

The exposure time, at which this condition can be achieved, is subject to the scatterers' concentration and size distribution as well as the camera's parameters and must be determined by the experimenter. In this research, milk samples with different fat content had different fat globules' concentration (and size distribution) and in consequence yielded speckle (images) of different brightness levels (and distribution). In the machine learning environment brightness is known to af-fect the performance of neural networks both positively and negatively [25,26]. In the case of our experiments we took care that the brightness levels are similar, by normalising each image brightness to 0–1 range. However, the brightness distribution (intensity histogram) could also vary due to image exposure time. Hence we investigated 4 exposure-level protocols using milk from Dairy 1 – 5 different lots per each Protocol. The protocols are described below, while their influence on the network performance is discussed in section 3.1.

| Fat Content [wt%] | Protocol exposure time [μs] | | |
|---|---|---|---|
| | A | B | C (D) |
| 0.5 | 0.6 | 4.0 | 4.5 (4.5) |
| | | | 4.0 (4.0) |
| | | | 3.5 (0.6) |
| 2.0 | 0.2 | 2.0 | 3.0 (3.0) |
| | | | 2.5 (2.0) |
| | | | 2.0 (0.2) |
| 3.2 | 0.1 | 1.0 | 1.5 (1.5) |
| | | | 1.0 (1.0) |
| | | | 0.5 (0.1) |

Table 1bis: Three exposure time protocols for recording speckle images. Protocol A: short exposure time, Protocol B: significantly longer exposure time and Protocol C: different exposure times within the same fat content recorded in proportion of 2:2:1 lots. Additional protocol *D* was composed from the three protocols in proportion of 2:2:1 lots.

Since longer exposure may influence the recorded shape of moving speckles, in Protocol A, we chose short exposure times. First we chose the shortest possible exposure for samples with 0.5% fat content, which were obviously the most transparent. We tried to obtain legible images with consistent intensity histograms with maxima as close to zero as possible (refer to Table 1bis). Consequently, the exposures for 2.0 and 3.2% fat contents were adjusted towards even shorter times, while their intensity histograms were monitored until the variation between sample types appeared minimal (refer to Figure 2, left panel). Though the frame brightness is always normalised in post-processing, such procedure is necessary since it may be well expected that the network may learn just the differences in brightness distribution between sample types, while we want it to learn rather finer details of the images. Figure 2 represents the intensity histograms of the first 10 consecutive frames of movies from each sample type for Protocol A (left panel) and Protocol B (right panel).

In Protocol B, an order of magnitude longer exposure time was chosen for 0.5% fat content, while the condition of not saturating the image was observed. Again, the exposures for 2.0 and 3.2% fat contents were adjusted until the variations in intensity histograms between sample types were smallest (refer to Figure 2, right panel). It is however plainly visible that the variation between intensity histograms both within one sample type and between sample types remained significant. The effect seems to be caused by speckle smear manifesting at larger camera accumulation times. In Protocol C, the images were recorded as follows: the first 2 lots with the first exposure, next 2 lots with the second exposure and the last lot with the third exposure (see Table 1 for the milk lots and Table 1bis for exposure times). Here we mimic the disturbances which may be encountered in the real-world applications, for instance the variation in brightness distribution caused by the automatic brightness control. The intensity histograms for Protocol C (not shown) are even more irregular than for Protocol B.
Finally, additional Protocol D was used as the cross-check for Protocol C – the lots from A, B and C protocols were mixed in 2:2:1 proportion respectively, where from the Protocol C only longest exposure times were used.

*2.5 Data extraction*

The recorded movies were extracted into 640×480 pixel RGB frames. Their green channels served as speckle images. In order to subtract a stationary background/immobile speckle – scattering on glass imperfections or immobile sample particles – averaging over each 200 consecutive images was performed and subtracted from each image during the extraction. Then the normalisation was performed, as stated earlier, and the intensity was expressed as *float* type, optimal for our GPU code. In Figure 3 (a) and (b) we present the normalised and background-subtracted speckle patterns from 0.5 and 3.2% fat contents respectively from Protocol B. Their corresponding mesh plots are presented in (c) and (d) respectively. It is visible that even for such higher exposure times, very limited pixels have maximum intensity value (compare histograms in Figure 2) – avoiding saturated regions being the first intuitive concern when selecting the exposure. Finally, it is worth emphasising that there is no known 'visual' technique for distinguishing between Figure 3 (a) and (b). Practically, the only physical property that may be differentiated here by the experimenter is the change in intensity/brightness for each sample type. However, it must be kept in mind that it was primarily adjusted by the camera's exposure time and is misleading.

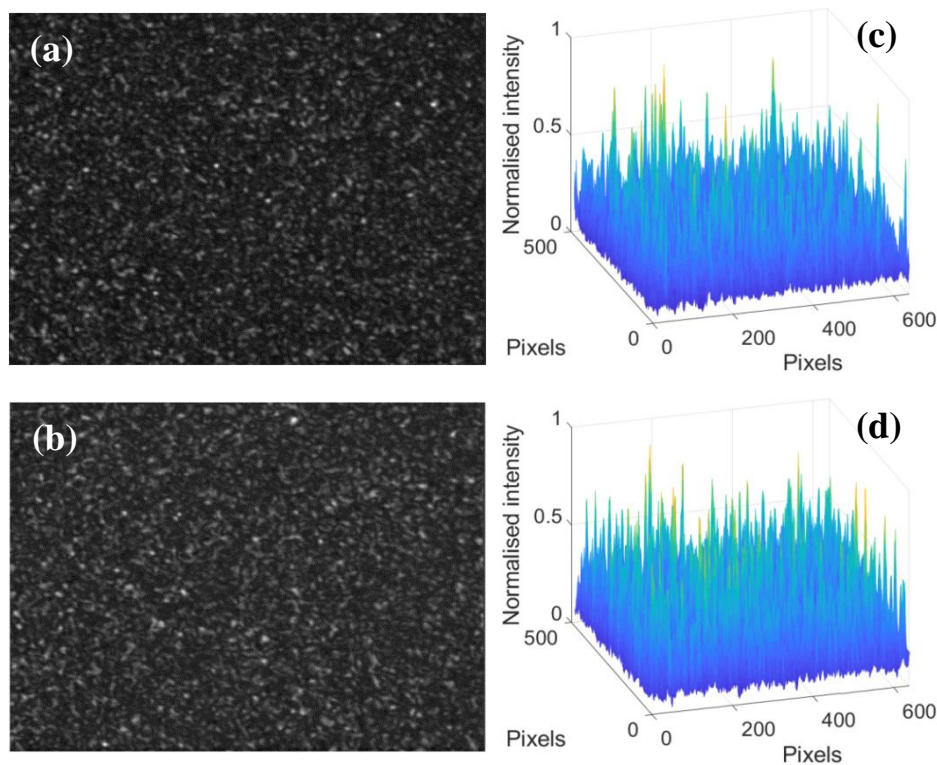

Figure 3: (a) and (b) are the normalised and background subtracted speckle patterns from 0.5 and 3.2% fat content respectively – the difference is hardly perceivable to a human observer; (e) and (f) are the corresponding mesh plots.

**2.6 *Convolutional neural network training***

The network used for training is the same as "*t7g24*" described in [25] and [28] however, we trained the network from scratch with a new set of weights. We also changed the output of the last fully connected layer to conform to our outputs. The frame extraction, image normalisation and network training were carried out in Matlab (2022b) environment on PC equipped with Tesla K40 GPU (nVIDIA). The network was trained on all images extracted from Version 1 with a randomised selection of 8:1:1 proportion for training, validation and test sets respectively. Each class corresponded to one fat content and was always constructed from images recorded from 5 different lots. Images from Version 2 served here as the independent set to test the network's ability to generalise. We trained the network using the two available optimization techniques [29] – first we applied adaptive moment estimation (ADAM) and then tried to refine the results with stochastic gradient descent with momentum (SGDM).

As mentioned above the network was first trained separately to recognise the milk fat content classes of Diary 1 in each protocol. In Figure 4 we present the training and validation loss per epoch during training of the networks. Figure 4 (a) and (b) represent training and validation loss of ADAM for Protocols A and B respectively. The training and validation loss of SGDM (not shown) appeared similar to (a) and (b) re-spectively, but not better. The cross entropy loss optimization during training of the networks for the protocols A and B (using the two training options) progressed steadily to the end of training. But at the end of training for the dataset recorded in Protocol C, there was a sudden fall of valida-tion accuracy for both training options, as can be seen in Figure 4 (c). This sudden jump (sometimes termed as the "variance shift") has been reported to be caused by Batch Normalisation (BN) and Dropout layers when both are used in the same network [30]. Also, for small mini-batch size, variance shift can result (even in absence of Dropout layers) from the difference between the batch normalisation layer statistics of the entire training data (computed at the end of training) and the mini-batches as reported in [31]. The latter could be possible in our case since we have used (only) multiple BN layers and a mini-batch size of 37 out of a training set of over 120,000 images in each protocol. After experiencing variance shift at the end of training of Protocol C, we cross-checked these results by applying Protocol D and suffered the same. While [31] suggested that increasing the mini-batch size might solve this problem, due to computer memory constraints we could not increase the mini-batch size beyond 37. Instead, we used "moving-average" to compute the BN statistics – an option available in newer versions of Matlab as an intended remedy for the issue [29]. Here we increased the validation patience as well as validation frequency (3 times per epoch) and trained for 10 epochs. The training loss is presented in Figure 4 (d). This attitude only apparently solved the problem. As we show in section 3.1, a problem with generalisation manifested then, which indicates the failure of the training.

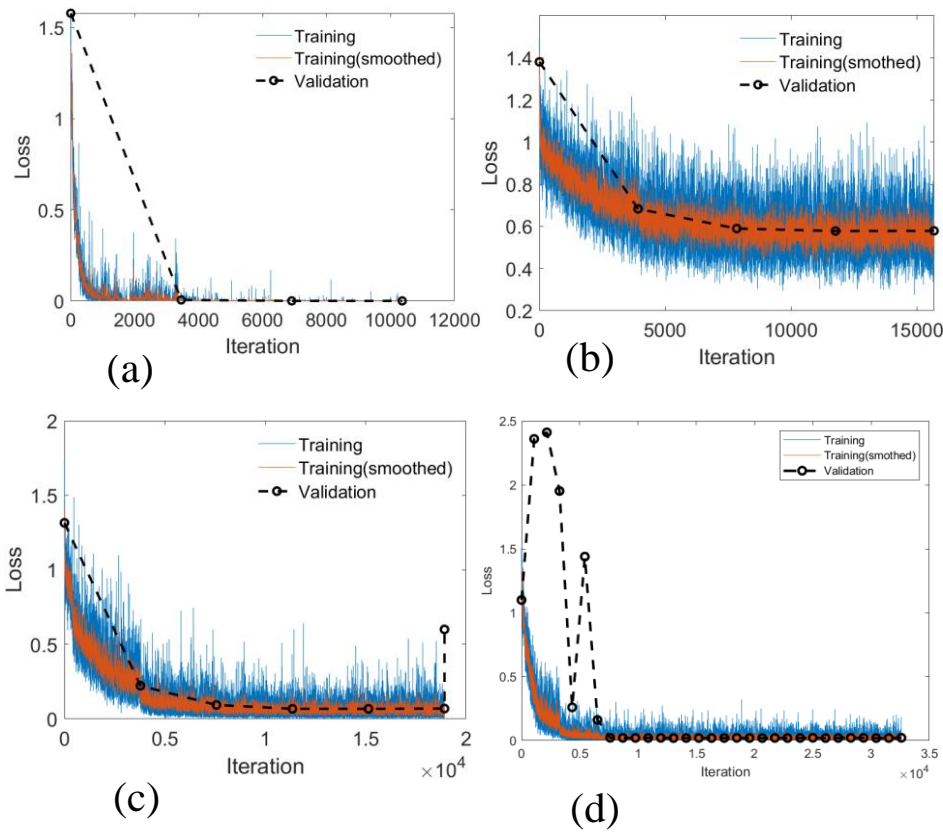

Figure 4: The loss function optimisation progress for training and validation subsets of Dairy 1 trained with ADAM optimizer. (a) Protocol A, (b) Protocol B, (c) Protocol C and (d) Protocol C using "moving average" instead of "population" option to compute the BN statistics. The abrupt jump at the end of training in (c) is due to so called variance shift, a common problem in network with batch normalisation layer. Epochs are marked with open circles in validation in (a) to (c), while in (d) the validation frequency was 3 times per epoch.

We can thus conclude that the variance shift in our case was caused by the different brightness distributions within the same class, since it did not occur in Protocols A and B. We report this phenomenon for the first time, as it seems. We found that the network learns or adapts to a specific brightness distribution (intensity histograms). Consequently, differences in distributions both within the same class and between classes must be carefully avoided. It can further be concluded that it is not enough to avoid brightness saturation, but the dynamic range of the image must be optimal. In other words, the speckle pattern must be fully developed and the individual speckles cannot be smeared over several sensor pixels. The results obtained with Protocols B and C support the notion that increase in exposure time gives the sensor ample time to accumulate passing speckles into several neighbouring pixels.

We found that the trained network from Dairy 1 performed poorly on data from Dairy 2, which also indicated insufficient generalisation. In consequence, we proceeded in two directions:
(i) We trained the network again from scratch with data from Dairy 2 and 3. However, in recording the data we used Protocol A and training was done using ADAM as expected to yield better results.
(ii) In order to tackle the problem of differences in milk from the dairy plants – the network can learn the features specific to the dairy plant – we trained the network on mixed dairies. Similarly, Version 2 of such images represents the independent set.

# 3. Results and Discussion

## 3.1 *Classification of fat content*

The classification confusion matrices for the test and independent sets of Dairy1 (using ADAM op-timizer for training) for Protocols A and B are presented in Figure 5, while in Figure 6 classification confusion matrices for Protocol C are presented. The top panel in Figure 6 represents the classification matrices when "population" option was used to compute the BN statistics, while the bottom panel represents the classification matrices when "moving-average" option was used. In both Figure 5 and 6, results for the test sets are shown in the left column, while for the independent sets – in the right. The independent sets contained 10 times more images than the test sets. When "population" (default) option was used to calculate BN statistics, in all the Protocols there are unambiguous classifications of milk samples into their appropriate fat content classes. However, it can be seen that as the exposure time increases, performances of the networks begins to decline.

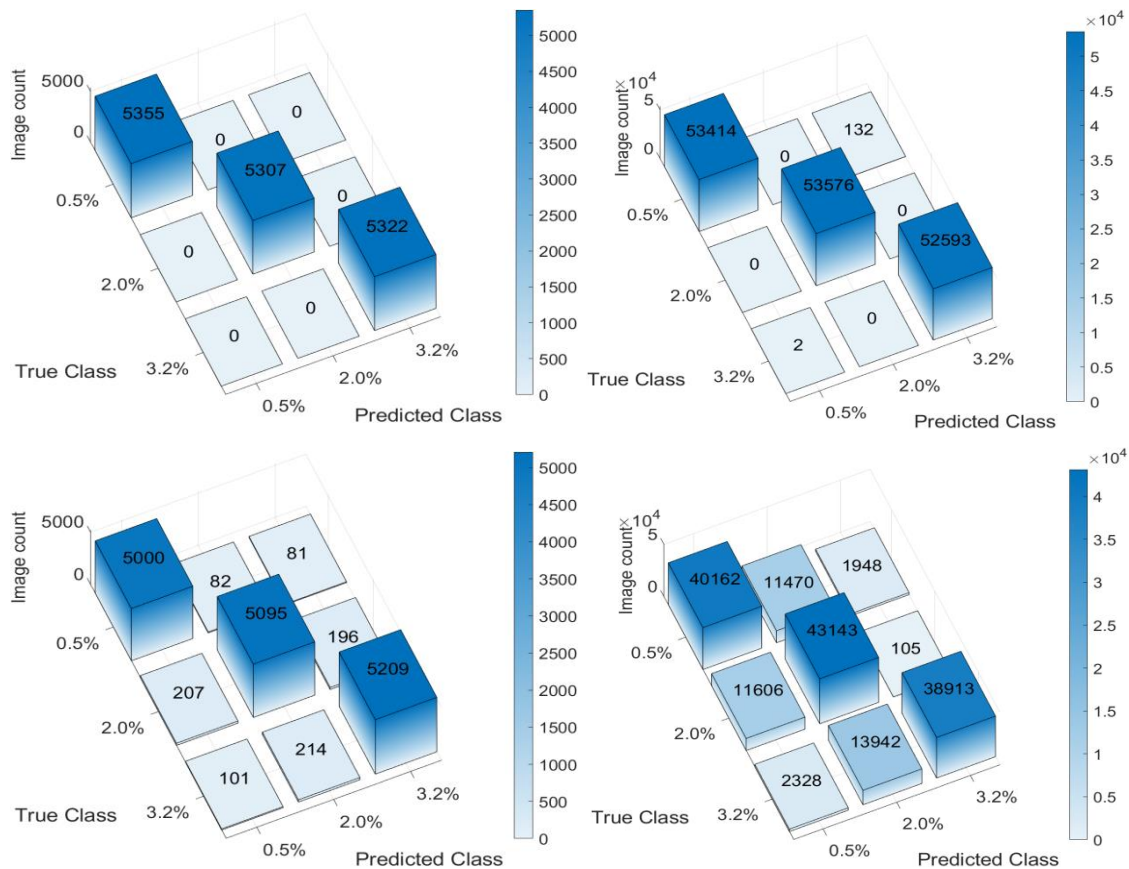

Figure 5: Classification confusion matrices of cow milk with 3 fat content classes produced by Dairy1. Results of training with ADAM optimizer for Protocol A (top panels) and Protocol B (middle panels). Results for the test sets are shown in the left column, while for the independent sets – in the right. In all training, 'population' was used to compute the batch normalisation statistics.

Using "moving-average" instead of "population" in Protocol C, produced rather misleading results. Though the variance shift at the end of training was eliminated, the performance was improved only for the test set, while for the independent set the results became ambiguous (Figure 6). Ultimately, it must be perceived as a failure of the attitude.

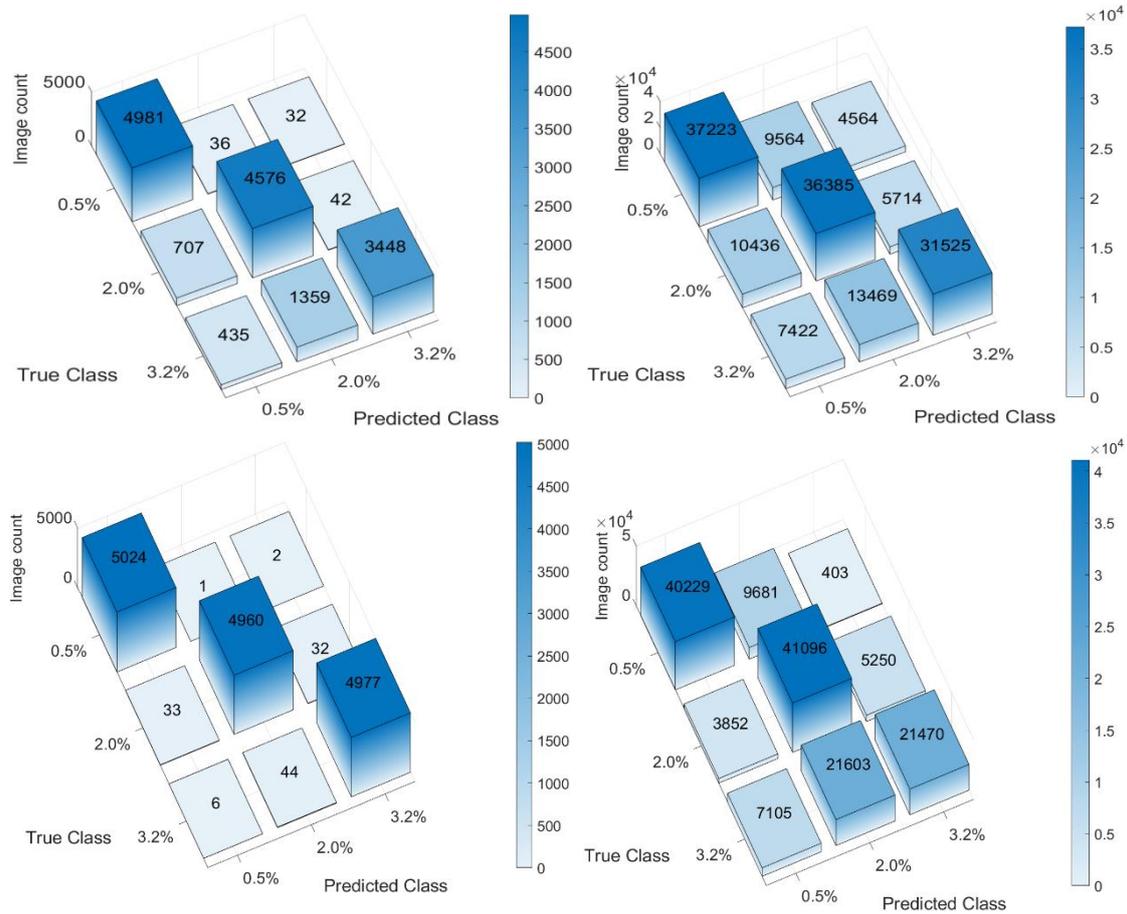

Figure 6: Classification confusion matrices of cow milk with 3 fat content classes produced by Dairy1 obtained from training with ADAM optimizer for Protocol C. Results of training with "population" and "moving-average" options used to compute the BN statistics are shown in top panels and bottom panels respectively. Results for the test sets are shown in the left column, while for the independent sets – in the right.

The classification confusion matrices of test and independent sets for Dairy 2 and 3, confirming the results obtained for Dairy 1, are presented in Figure 7. The training set, in the case of Dairy 3, consisted of 4 classes, one of which (2%) corresponded to "fresh" milk – in contrast to UHT. It is worth highlighting that with the addition of 1.5% fat content class, the fat content difference between 1.5 and 2.0% is relatively small. However, the neighbouring 2% class also differs in terms of the fresh/UHT parameter. Thus, the ability of the network to distinguish fat content difference down to 0.5% still had to be verified. Ultimately, we expected the classification to be regardless of dairy plant (processing differences) or milk type (fresh/UHT). In view of that, the most demanding training set – labelled as Mixed Diary – was constructed (see Table 1). It consisted of four fat content classes mixed from 5 dairies (3 dairies in each class) and 2 types of milk.

Two independent mixed sets were constructed: (i) the set consisting of Versions 2 from the lots used in the training set, (ii) the set consisting of the extra lots – not used for training. The results of classification of milk for Mixed Dairy are presented in Figures 8 and 9 (extra lots). Comparing the result for extra lots to the independent set (Version 2 of the lots used in training) assures us that when all irrelevant features in dairies are suppressed by randomisation in a larger training set and only relevant features in milk are fully represented, the technique will classify milk samples into their appropriate fat contents.

Hence, the low recognition of 1.5% fat content could be attributed to the fewer representations of Dairies 4 and 5 in training, while for 2.0% we expect that the fresh/UHT milk types (different processing) mix was not sufficiently balanced. We expect that it would be beneficial if the images were generated from milk produced by several dozen dairy plants, optimally over a period of a year, which would supress learning irrelevant features. A summary of the accuracies for validation, test and independent sets for all training and testing using ADAM for Diary 1 for all protocols are presented side-by-side in Tables 3. Also presented in the table are the results for Dairy 2, 3 and Mixed Dairy. The accuracy obtained from the classification of extra lots from Mixed Dairy is presented in bold text.

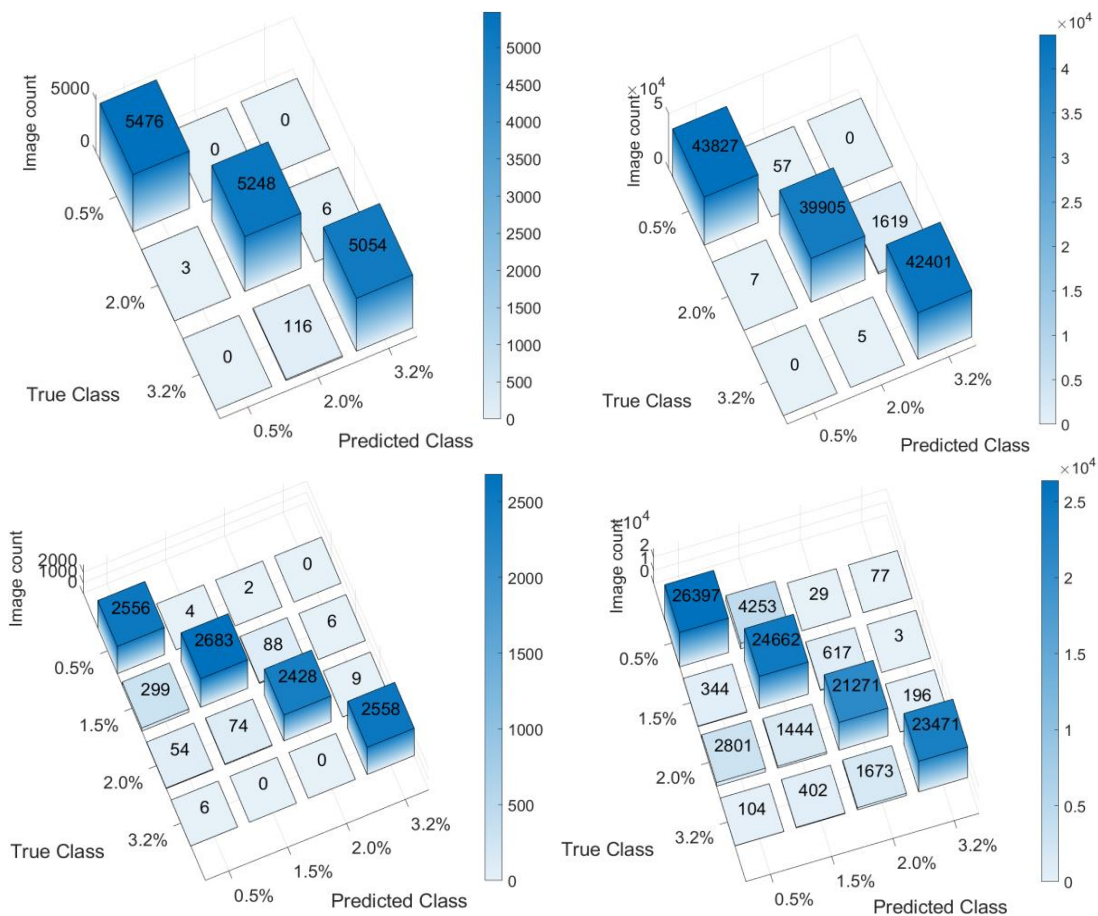

Figure 7: Classification confusion matrix of milk fat content produced by Dairy 2 (upper row) and Dairy 3 (lower row); test sets (left panel) and independent sets (right panel).

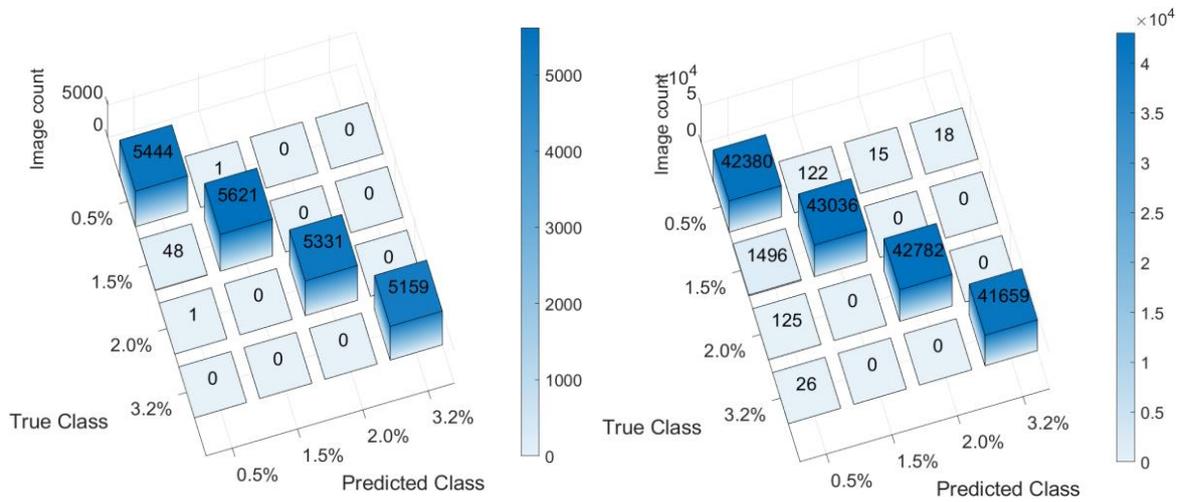

Figure 8: Classification confusion matrix of speckle images from Mixed Dairy; test set (left panel) and independent set (right panel). Mixed Dairy was created by mixing four fat content classes and 2 milk types from five dairies (see Table 1).

### 3.2 *Temporal correlations in image sequence*

Since we record a sequence of images, it could be possible to exploit also the temporal correlation between the images – if they exist – for network training. Our frame rate is several orders of magnitude lower than that used in DLS technique, but studying the speckle field along the temporal dimension might yield even finer suspension (milk) parameters. So far, such attitude has significantly exceeded our hardware capabilities but we wanted to verify its feasibility. The temporal correlation analysis performed on the image sequences that we recorded, revealed that correlation exists but only between the first few successive images (see Figure 10 – the highest correlation of 1 is the autocorrelation of the first image in the movie). It seems that the analysis of this simple correlation only, cannot be used to estimate, for instance, the fat content of milk, but could be used in future together with the spatial image analysis.

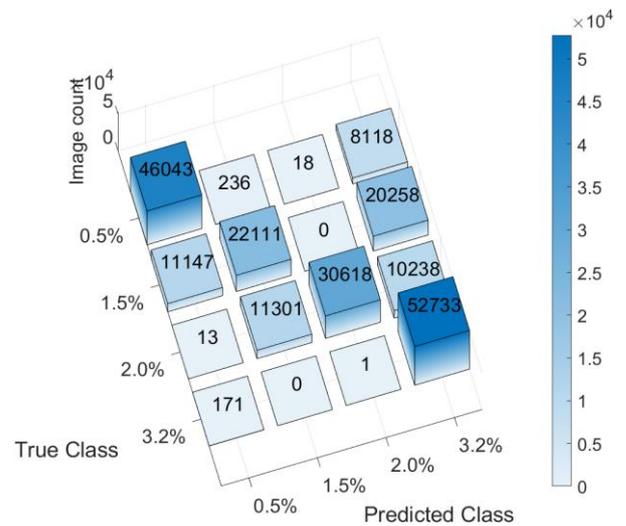

Figure 9: Classification confusion matrix of speckle images from the extra lots from the Mixed Dairy (see Table 1; compare Figure 8).

### 4. *Conclusion*

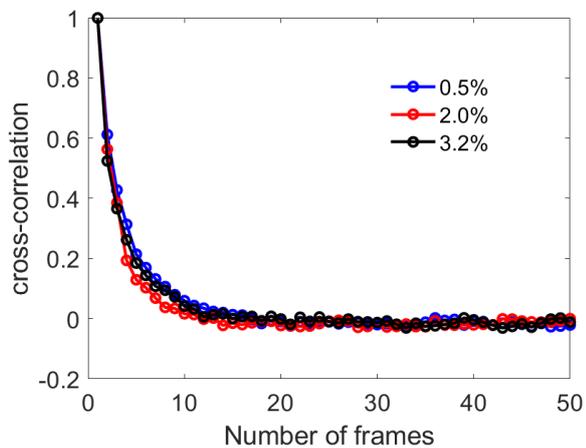

Figure 10: Cross-correlation between the first frame and successive frames in a single movie for three fat content classes. The highest correlation of 1 is naturally the autocorrelation of the first frame.

Non-destructive, cheap and fast – preferably online – measurements of selected properties of milk, as well as other real-life suspensions, are of vivid interest. Accurate measurement of such parameters as total fat are important for assessing milk production quality. We have investigated the potential of combining machine learning and speckle photography to directly link speckle patterns to four fat content classes. We anticipated differences in speckle pattern to be directly influenced by concentration and size distribution of fat globules. However, milk is a multi-component real-life suspension and its components are not entirely separable without influencing others. There are also seasonal variations in milk properties. This posed a risk that the milk parameters could not be classified separately. However, our experiments showed that such classification is feasible. We discovered that when the speckle is recorded such that the intensity histograms are kept similar for all images (falling into all the distinguished classes) and most probable intensity in the pattern is close to zero (shorter exposure times and the same within the class), the convolutional neural network yields unambiguous classifications. Our technique requires only 10 μL of undiluted milk sample in a thin cuvette and uses a very simple experimental setup for recording movies. Once the network is properly trained, the whole process of recording several frames, extraction, normalisation and classification can be deemed practically on-line.

| Data | | Validation accuracy % | Test set accuracy % | Independent set accuracy % | |
|---|---|---|---|---|---|
| Dairy 1 | Protocol A | 100.00 | 100.00 | 99.92 | |
| | Protocol B | 94.44 | 94.56 | 74.70 | |
| | Protocol C | 83.50 (99.19) | 83.28 (99.22) | 67.26 (68.22) | |
| Dairy 2 | | 99.57 | 95.19 | 98.92 | |
| Dairy 3 | | 94.54 | 94.97 | 88.92 | |
| Mixed Dairy | | 99.74 | 99.77 | 98.95 | **71.13** |

Table 3: A summary of validation and testing accuracies for all training with ADAM optimizer. The accuracy for classifying extra lots irrespective of dairy is presented in bold text. Results pertaining to "moving average" option are shown in brackets.

## 5. Acknowledgements

This research was funded in whole or in part by National Science Centre, Poland, grant 2021/41/B/ST3/00069. For the purpose of Open Access, the author has applied a CC-BY public copyright licence to any Author Accepted Manuscript (AAM) version arising from this submission

The authors would like to thank Tomasz Jakubczyk and Gennadiy Derkachov for consultations.